\documentclass[conference]{IEEEtran}
\IEEEoverridecommandlockouts
\usepackage{cite}
\usepackage{amsmath,amssymb,amsfonts}
\usepackage{algorithmic}
\usepackage{graphicx}
\usepackage{cite}
\usepackage{textcomp}
\usepackage{gensymb}
\usepackage{xcolor}
\usepackage{setspace}
\usepackage{lipsum}
\def\BibTeX{{\rm B\kern-.05em{\sc i\kern-.025em b}\kern-.08em
    T\kern-.1667em\lower.7ex\hbox{E}\kern-.125emX}}

\begin{document}
\bstctlcite{IEEEexample:BSTcontrol}

\title{Towards Modular and Accessible AUV Systems\\
    \thanks{This work is supported by the National Science Foundation (NSF) under the award \#2154901.}
    \thanks{$^{1}$Graduate School of Oceanography, University of Rhode Island, Narragansett, RI 02882, USA. Email:{\tt\small mzhou@uri.edu}}
    \thanks{$^{2}$Department of Ocean Engineering, University of Rhode Island, Narragansett, RI 02882, USA. Email:{\tt\small \{farhang.naderi, tony.jacob, linzhao, yweifu\}@uri.edu}}
    \thanks{$^3$Department of Mechanical, Industrial, and Systems Engineering, University of Rhode Island, Kingston, RI 02881, USA. Email:{\tt\small \{wmcconnell, cyuan\}@uri.edu}}
    \thanks{$^4$ Department of Electrical and Computer Engineering, Stevens Institute of Technology, Hoboken, NJ 07030, USA. Email: {\tt\small mpanjnan@stevens.edu}}
    \thanks{$^5$Department of Marine Technology, Norwegian University of Science and Technology, Norway. Email: {\tt\small emir.cem.gezer@ntnu.no}}
}

\author{Mingxi Zhou$^{1}$, Farhang Naderi$^{2}$, Yuewei Fu$^{2}$, Tony Jacob$^{2}$,  Lin Zhao$^{2}$,\\
Manavi Panjnani$^{4}$, Chengzhi Yuan$^{3}$, William McConnell$^{3}$, Emir Cem Gezer$^{5}$ }
\maketitle

\begin{abstract}
This paper reports the development of a new open-access modular framework, called Marine Vehicle Packages (MVP), for Autonomous Underwater Vehicles.
The framework consists of both software and hardware designs allowing easy construction of AUV for research with increased customizability and sufficient payload capacity.
This paper will present the scalable hardware system design and the modular software design architecture.
New features, such as articulated thruster integration and high-level Graphic User Interface will be discussed.
Both simulation and field experiments results are shown to highlight the performance and compatibility of the MVP.
\end{abstract}

\section{Introduction}
Autonomous underwater vehicle is a growing area since they are great tools for ocean research and defense purposes.
Commercial-off-the-shelf (COTS) AUVs are supplied with proprietary software are great when they are used as an equipment for collecting scientific data, e.g., survey the seabed and profile the water column.
However, these COTS AUV platforms may be challenging for researchers in academia to integrate and test sensors and advanced autonomy algorithms due to high upfront cost (in the order of \$100k) for a small to medium size AUV and the low customizability in software.
To this end, low-cost AUVs and open-access platforms has been researched.
For example, LoCO AUV \cite{Edge2020DesignAE} was designed and developed for underwater computer vision research, Hippo AUV \cite{Duecker2020HippoCampusXA} was constructed with excellent manuverabiltiy, and miniROV AVEXIS \cite{Griffiths2016AVEXISAquaVE} is prototyped.
Indoor experiments have been conducted on these small portable AUVs with emphasize on computer vision and controls.
Recently, a wide range of micro-AUVs have been developed and field tested (\cite{Rypkema2023PerseusAT, Randeni2022MorpheusAA}) with new features in maneuvering, localization, and autonomy.
The small and affordable features making them great candidates for swarm and cooperative operations.
However, for scientific mission, e.g., mapping or water column profiling, a suite of payload sensors are required to collect multi-modal data sets.
Therefore, the micro-AUV may run out of space for user demands.
Blue ROV is another potential option for underwater research, but they are mainly designed for remote operation which offers limited high-level autonomy (e.g., mission control and finite state machine) for autonomous mission.
Moreover, their control system is tuned and configured for the specific thruster placements, limiting actuation customization for better manuverability or efficiency.

On the software aspect, to our best knowledge there is only few AUV Guidance Navigation and Control frameworks.
MOOS-IvP \cite{Benjamin2010NestedAF}, the most notable marine robotic framework, has enabled a wide range of research projects from adaptive sampling to multi-AUV operations.
It comprises two main components: the MOOS for inter-process communication (IPC), and IvP Helm for guidance and autonomy. 
User could easily implement custom software and interface with the core systems using the Mission Oriented Operating Suite (MOOS).
However, MOOS-IvP only contains high-level behaviors that assumes the platforms has already established its pose control and localization systems.
Moreover, when user trying to leverage the resources available in Robotic Operating Systems (ROS) (one of the most used middleware in robotics), user will require to maintain both ROS and MOOS middleware and the bridge between them on the vehicle.
In ROS environment, COLA2 \cite{Palomeras2012COLA2AC} is a ROS-based GNC systems for AUVs.
But it is a licensed software that are only available on the platforms (e.g., Sparus II and Girona 500) from IQUA Robotics.

To this end, this paper presents the progress in developing an open AUV hardware and software framework, called Marine Vehicle Packages (MVP), beyond the preliminary results from our previous paper \cite{Gezer2022WorkingTT}.
This development seeks to fill the gap in small-medium size open-accessible AUV platforms and ROS-based GNC systems that could provide sufficient payload capacity and scientific mission capability at the affordable level (about \$20k per AUV).
By providing open-source access software \cite{mvp_readme} and affordable hardware, the authors seek to lower the barrier-of-entry for AUV research, especially the development of Guidance Navigation and Control systems, and offer a customizable framework for high-level autonomy and advanced algorithm development.

The technical contribution of this paper is two-fold.
First, we will provide detailed information on the hardware design (Section II) which allows for AUV customization with expanded sensor payloads.
Second, we will present the new MVP software features in Section III, including a GUI system and the generalized pose control system for articulated thrusters.
To demonstrate the compatibility of the software, we will provide results from our field tests and simulation tests on a variety of marine robotic platforms in Section IV.

\section{MVP framework}

\begin{figure}[h]
    \centering
    \includegraphics[width=1.0\linewidth,trim=0cm 1cm 0cm 0cm]{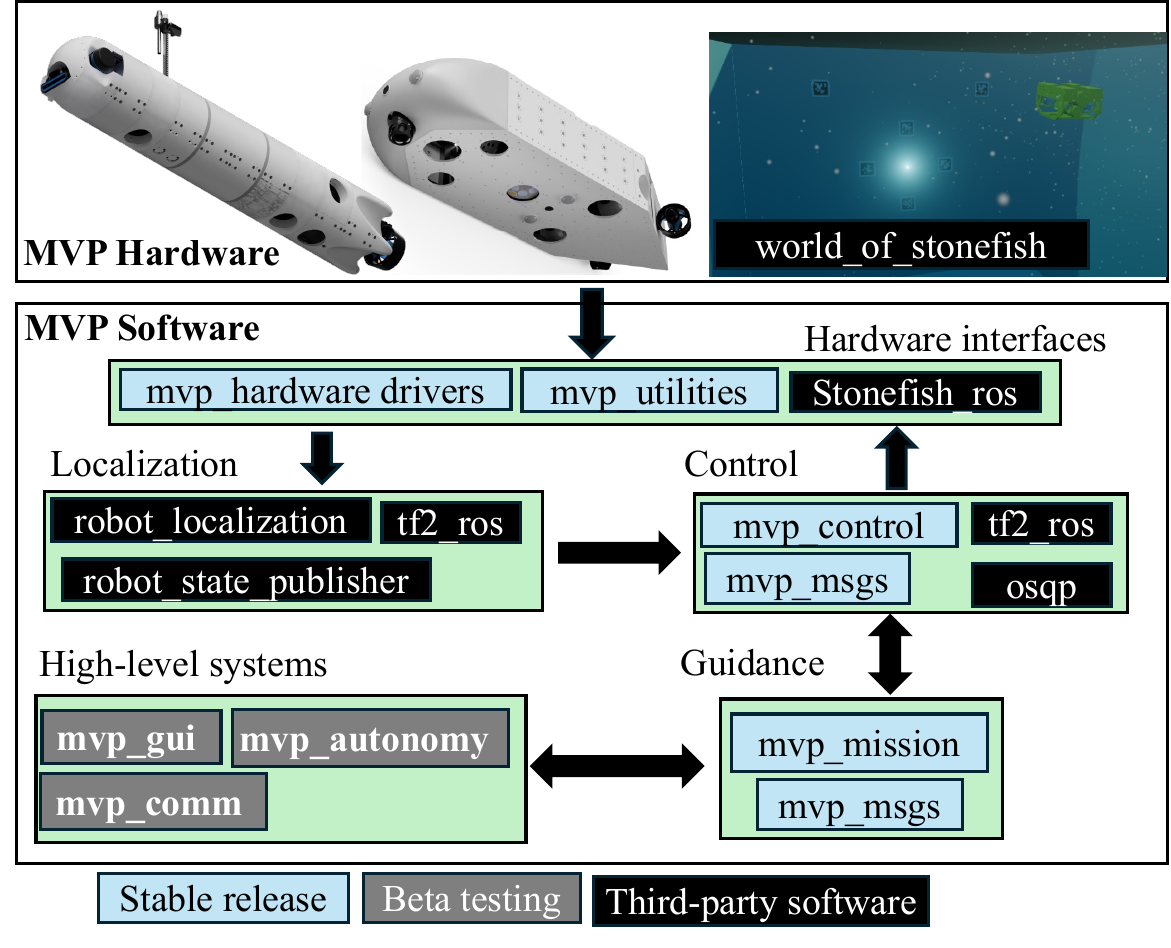}
    \caption{MVP framework system diagram with readme in \cite{mvp_readme}}
    \label{fig:mvp_overview}
\end{figure}

In Fig. \ref{fig:mvp_overview}, we present the overview of the MVP framework which consists of the hardware and software portion.
The MVP hardware mainly includes a sets of Printed Circuit Boards (PCBs) and other interface boards that can be packaged inside a small pressure housing, while the MVP software is a suite of ROS-packages developed for hardware interfacing, localization, control, guidance, and high-level decision systems.
We also leverage the Stonefish simulator \cite{Cielak2019StonefishAA} for software development and debugging.

\vspace{-1ex}
\vspace{-2ex}

\subsection{MVP hardware}
\begin{figure}[h]
    \centering
    \includegraphics[width=1.0\linewidth,trim=0cm 1cm 0cm 1cm]{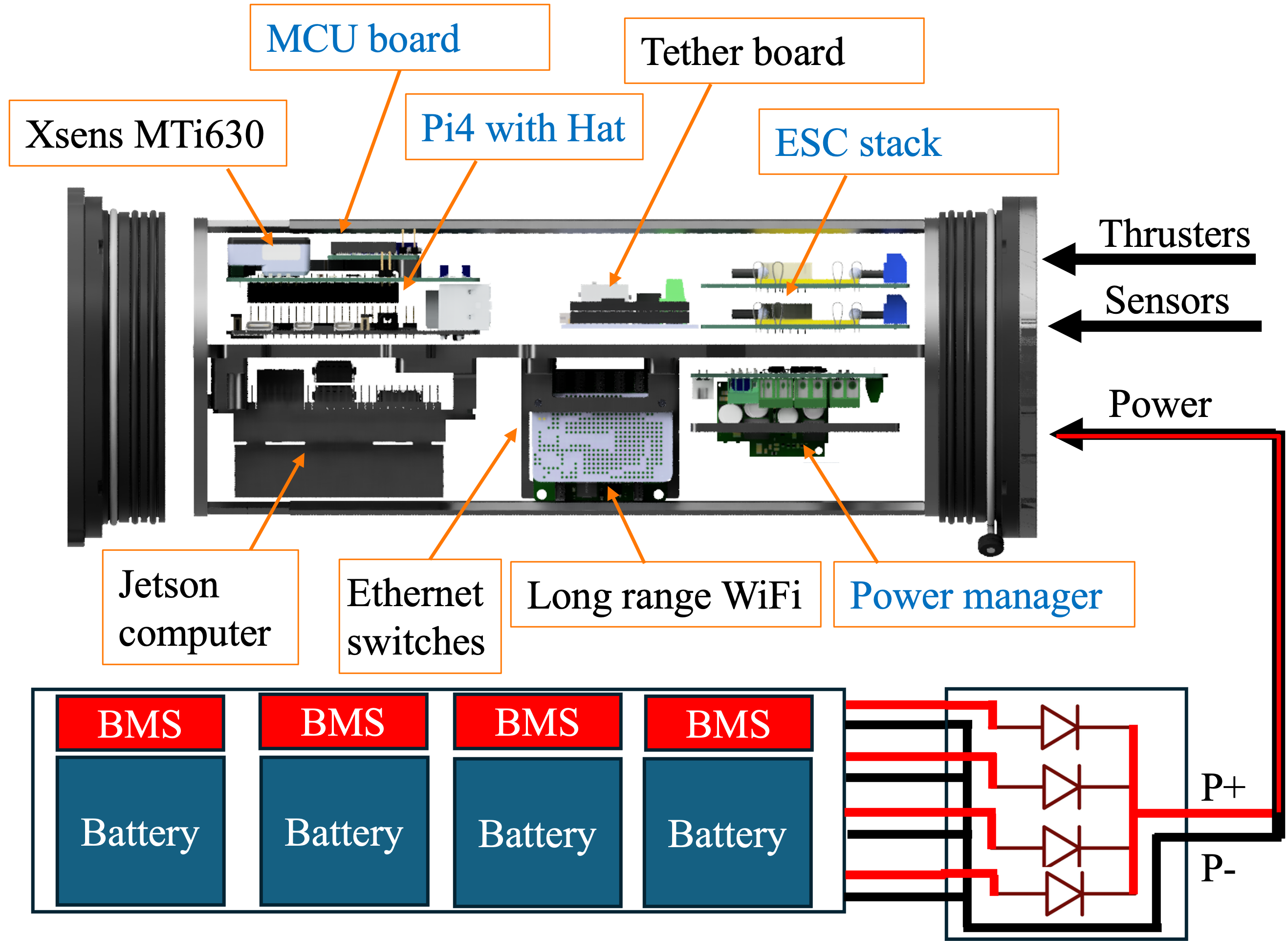}
    \caption{Core electronics diagram arranged in two pressure vessels}
    \label{fig:electronics}
\end{figure}

Compared to our previous prototype, we have minimized our electronics footprint with customized PCBs made for easy sensor expansion.
Figure \ref{fig:electronics} presents the interior arrangement of the electronics housing (300 mm long 4-in inner diameter Blue Robotics pressure housing) and the battery housing (400 mm long 4-in inner diameter).

Inside the electronics pressure vessel, we have three customized PCBs.
First, we have designed a power manager board to distribute the power from battery voltage to different voltage levels (5v, 15v, and battery voltage) for different sensors and components.
For power monitoring purpose, the board also contains a current and voltage sensing circuit which outputs analog voltage signals.
Three 8-Amp rated MOSFETs (Onsemi FDS8884) are integrated on the board such that the power of sensor payloads (e.g., DVL, sonar, and backseat computer) could be controlled using digital IOs from a single-board computer.
There are two computers inside the pressure vessel.
The PI-4 (front-seat computer) is used for running the guidance, navigation and control systems and basic sensor drivers for IMU, DVL, and etcs.
The customized PCB hat is designed for PI-4 for system expansion with two I2C buses and 8 RS232 ports.
The board will be compatible with other SBCs having the same PI GPIO layout.
In contrast, the Jetson embedded computer is used to perform computational demanding programs, such as camera and imaging sonar drivers, sensor-driven path planner, and computer vision algorithms.
The microcontroller (MCU) board is designed based on RP2040 for controlling motors (thrusters, and servos). 
It interfaces with the PI-4 using NMEA strings, and safety timeout is implemented to stop motors when PI-4 is stalled.
For sensor payload expansion, we have included two 5 ports Gigbytes Ethernet switches (Botblox), and have 8 RS232 ports on the SBC hat.
We have reserved connectors on the endcap for both USB and 10/100Mbps device expansion.
For communication, MVP system has a single-twist pair tether board and a long-range Wifi module (Doodlabs 900MHz/2.4GHz dual frequency modem).
The total power consumption of the electronics system (including the backseat computer) is about 15-18 Watts.
All the designs are available  in \cite{mvp_readme}.

For battery management, we have designed a diode array (30A per channel) PCB allowing for paralleling up to 4 battery packs.
The diode array circuit is encapsulated using high thermal conductivity epoxy and placed outside the pressure housing, allowing users to separate the battery housing into different housings when needed.
We also have installed a Battery management system (BMS) for each battery pack to protect from over discharge and over current situations.
The battery packs can be recharged independently through the bulkhead connectors on the endcap without opening the pressure housing (4-inch 400 mm Blue Robotics pressure enclosure).
We recommend using four 18V-13.8Ah Li-on battery pack from MaxAmps, resulting in about 1kW hour energy capability.
The overall endurance of the AUV systems is estimated to be about 2 (30A current draw) to 10 hours (moderate payload).

\subsection{MVP Software}
As illustrated in Fig. \ref{fig:mvp_overview}, the MVP software consists of several blocks. 
The hardware interfaces block stores the scripts that are needed to interface with sensors and simulations. 
The scripts converts sensor readings or simulated sensor data into standard ROS messages, allowing for easy adaption of other existing ROS packages.
For example, MVP uses the Extended Kalman Filter in Robot Localization package \cite{MooreStouchKeneralizedEkf2014} for generating a baseline AUV odometry using the measurements from GPS, DVL, and IMU. 
To minimize localization errors, we also leveraged tf2\_ros packages to correct for any impacts caused by sensor mounting offsets.
From our field experiments, the underwater position drift rate is about 5\% when combining DVL and IMU.
    
In the control block, the mvp\_control compares the vehicle pose from localization block and the desired pose from guidance block to generate control commands to the hardware interfaces. 
The controller is implemented using thruster allocation method and quadratic programming, as discussed in our previous paper\cite{Gezer2022WorkingTT}.
For easy use, we have leveraged the tf2\_ros to automate the thruster allocation matrix generation process.

As shown in Eq. \ref{eq:allocation_one_thruster}, a unit force $\boldsymbol{T} = [1\ 0 \ 0]^T$ from a thruster frame, can be converted into the torques and forces in body frame $\{b\}$ (first 6 rows) and earth frame $\{e\}$ (last 6 rows) based on the rotation matrix $\boldsymbol{R}^b_k$ between the $k$-th thruster frame $\{k\}$, mounting offset $\boldsymbol{r}^b_k$ in the body frame, and the rotation matrices ($\boldsymbol{R}^e_b$ and $\boldsymbol{J}^e_b$) between body frame and the earth frame (e.g., East-North-Up or North-East-Down frame) derived using the current Euler angles.
We then concatenate the matrix from each thruster horizontally to form the thruster allocation matrix for an AUV and update it iteratively for the Euler angle changes or even position changes, as shown in Eq. \ref{eq:allocation_all} where the expected force and torques generated from the thrusters is $\boldsymbol{\tau}$ and $\boldsymbol{F}$ is an $N$ by 1 matrix where N is the number of thrusters.
By applying quadratic programming with force constraints for individual thrusters, we could solve $\boldsymbol{F}$ to minimize $\boldsymbol{\tau}$ and the requested force and torques ($\boldsymbol{\tau}^*$) from the controller.
The thruster commands are then solved based on the thrust-command polynomial equation fitted using manufacturer data.
Worth noting that, our PID controller system allows users to define different controller modes with different controlled degrees of freedom (DOFs).
Since some states will be ignored, the thruster allocation matrix will be resized and the PID controller gains will be updated when the control mode has been changed.

\vspace{-1ex}
\begin{equation}
\label{eq:allocation_one_thruster}
    \bold{M}_k = 
    \left[\begin{array}{c}
         \boldsymbol{R}^b_k \boldsymbol{T}\\
         \boldsymbol{r}^b_k \times (\boldsymbol{R}^b_k \boldsymbol{T})\\
         \boldsymbol{R}^e_b \boldsymbol{R}^b_k \boldsymbol{T}\\
         \boldsymbol{J}^e_b (\boldsymbol{r}^b_k \times (\boldsymbol{R}^b_k \boldsymbol{T}))
    \end{array}\right]
\end{equation}
\begin{equation}
    \label{eq:allocation_all}
    \boldsymbol{\tau} = \boldsymbol{M}\boldsymbol{F}
\end{equation}
\begin{equation}
    \label{eq:qp_solver}
    \begin{array}{rl}
    \boldsymbol{F} = & argmin((\boldsymbol{\tau}^*-\boldsymbol{\tau})^T (\boldsymbol{\tau}^*-\boldsymbol{\tau}))\\
    & s.t. \boldsymbol{A F}\leq \boldsymbol{B}
    \end{array}
\end{equation}
\begin{equation}
    \label{eq:poly}
    F = \sum^n_{i=0} a_i u^i
\end{equation}

The desired pose is generated from our guidance block in which a finite-state-machine (FSM) is configured.
In each state, the user could attach one control mode and multiple behaviors (e.g., path following, tele-operation, and surfacing) 
All attached behaviors will generate desired poses based on the preprogrammed logic but have to be configured at different priority in a state.
For instance, the path following behavior will output desired heading and depth to following a list of 3D waypoints, and periodical surfacing will output desired depth when a predefined time interval has reached.
If two behavior is commanding the desired values in the same DOF the desired value from the behavior with a higher priority will be selected.
Since our behaviors are created as plugins, user could create multiple instances of a behavior with different parameters and assign to different states. 
For example, different surfacing interval can be configured in two different survey states.

\vspace{-2ex}
\begin{figure}[h]
    \centering
    \includegraphics[width=1.0\linewidth,trim=0cm 1cm 0cm 0cm]{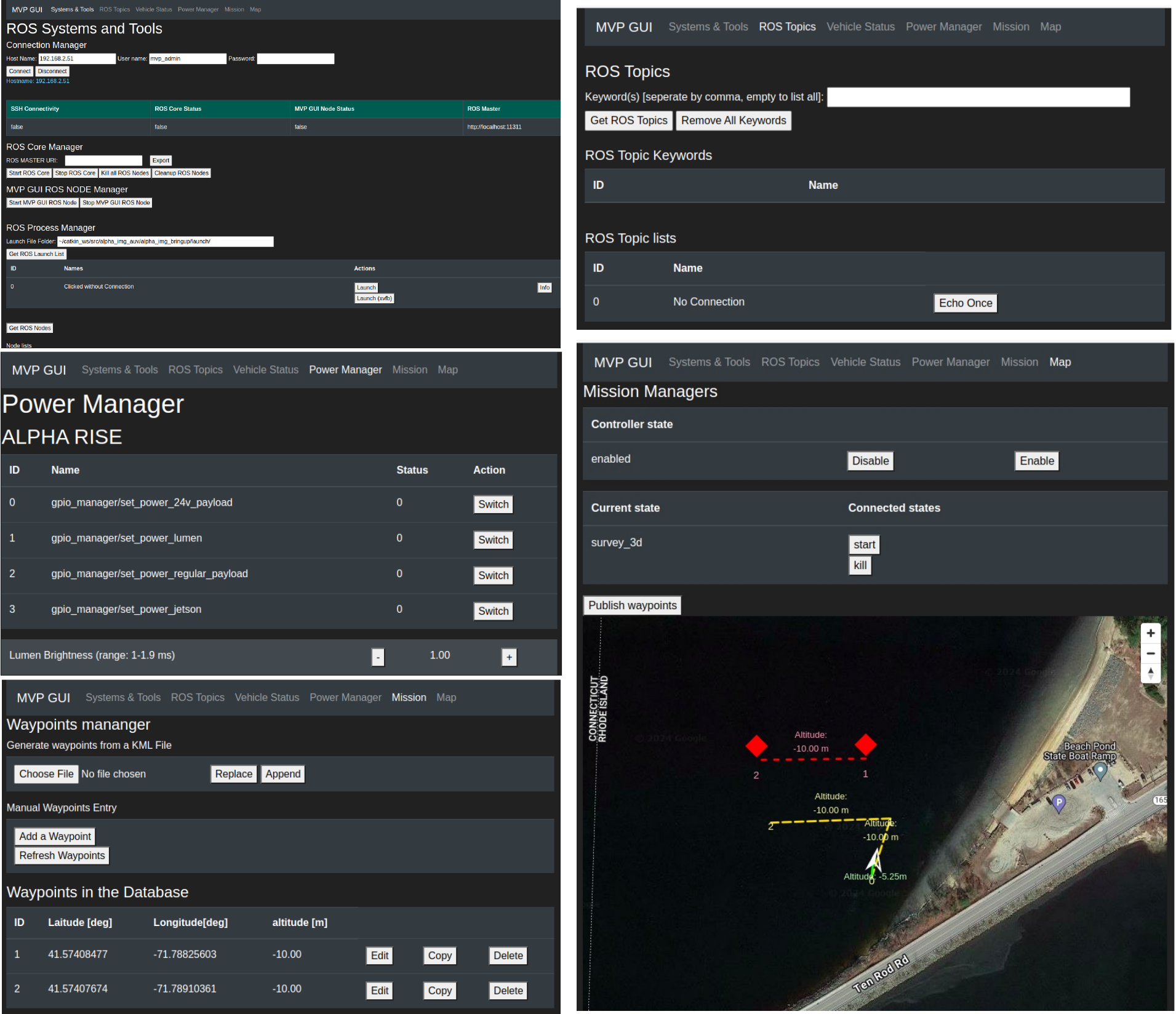}
    \caption{Screenshots of the MVP GUI system}
    \label{fig:mvp_gui}
\end{figure}
\vspace{-1ex}

To increase user experience, the high-level systems will create objectives (e.g., waypoint lists and state changes) for the guidance systems based on user actions or sensor outputs.
For instance, the mvp\_autonomy contains path planners that will generate waypoints for the guidance system based sonar and acoustic systems for collision avoidance, coverage seafloor mapping, and AUV following. 
We also have created a web-based GUI system (mvp\_gui) to allow AUV operation without terminal commands.
The GUI consists of a total of 6 pages (vehicle status page is not shown), as shown in Fig.\ref{fig:mvp_gui}.
The vehicle status page (not shown) shows vital information of an AUV, e.g., pose and voltage.
The Systems\&Tools page allows users to ssh into the robot or other machines in the same network to start launch files.
Once, the launch file has been started, user could also check the existing ROS nodes, and also obtain a most recent message from a ROS topics page where keyword filter can be applied.
The power manager page enables the user control on MOSFETs on the SBC to turn on/off sensors and payloads.
Mission and Map pages are designed for running AUV missions.
On the Mission page, user could manually enter and edit the GPS coordinates and altitudes of a list of waypoints or create waypoints from a KML file. 
In contrast, on the Map page, user can drag the waypoints on an interactive map where AUV's current position and past 20 positions are visible.
Besides that, the Map page also allowing user to enable or disable controllers and change the vehicle states by clicking the tabs.

Meanwhile, we are developing the mvp\_acommm package which allows AUV mission update and monitoring using acoustic modems.
The package leveraged the DCCL and dynamic buffer features in the Goby software \cite{goby} to compress acoustic messages and to manage the time slots allowing for the future expansion of multi-AUV operations.

\subsection{Articulated thruster control}
The major upgrade from our previous work \cite{Gezer2022WorkingTT} is the integration of 1-DOF articulated thrusters into the mvp\_control.
To avoid increased computational demand ($2^n$ solutions) when solving the allocation problem with multiple articulated thrusters, we have designed a new approach to simplify the problem while allowing the articulated thrusters to be independently angled.
Unlike existing works where they solve the thruster forces in x and y axis in a static frame \cite{Fossen2006ASO}, e.g., the thruster base frame, or they constrain the thruster to have the same tilting angle \cite{Jin2015SixDegreeofFreedomHC}, our approach defines the forces in the thruster frame (a rotating frame) and allowing the thruster to be independently controlled.

\vspace{-3ex}
\begin{figure}[h]
    \centering
    \includegraphics[width=1.0\linewidth,trim=0cm 1cm 0cm 0cm]{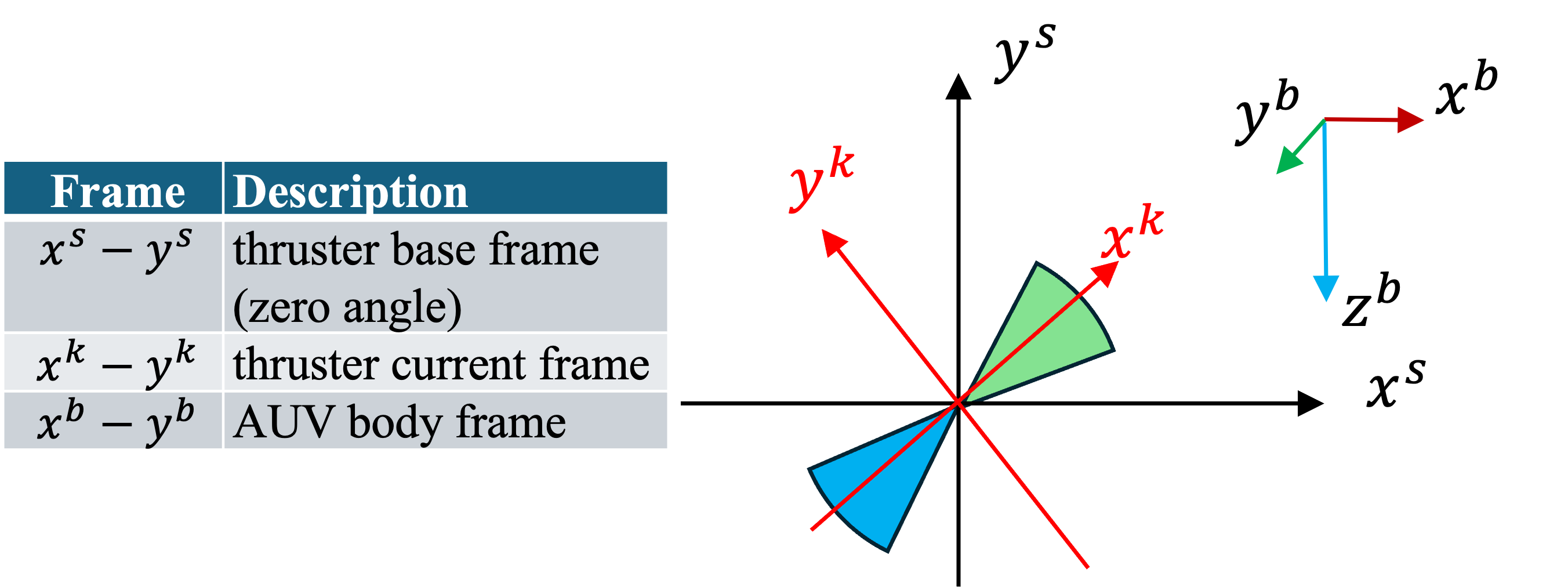}
    \caption{A diagram of coordinate systems defined for articulated thrusters.}
    \label{fig:thruster_diagram}
\end{figure}

\vspace{-1ex}
\begin{equation}
    \label{eq:thruster_allocation}
        \boldsymbol{\tau}_k
    = 
    \boldsymbol{M}_k
    \left[\begin{array}{c}
        X^k \\
        Y^k 
    \end{array}\right]
\end{equation}

\begin{equation}
\label{eq:constraints}
    \begin{array}{c}
        0\leq X^k\leq T_{Max}\\
        \tan(\omega \delta t) X^k + Y^k >0\\
        -\tan(\omega \delta t) X^k + Y^k <0
    \end{array}
\end{equation}
\begin{equation}
    \label{eq:solution}
    F^k = 
    \sqrt{X^k*X^k + Y^k*Y^k}
\end{equation}
\begin{equation}
    \label{eq:solution2}
    \Delta\alpha^k = \tan^{-1}(Y^k/X^k)
\end{equation}

Figure \ref{fig:thruster_diagram} shows a diagram of the coordinate systems we defined for articulated thrusters.
In the defined coordinate system, the force generated by the thruster in the next time step can be divided into two sub-components along the x and y axis in the current thruster frame ($x^k-y^k$).
Their contribution can then be expressed using Eq. \ref{eq:thruster_allocation}, where $\boldsymbol{M}_k$ maps the two forces ($X^k$ and $Y^k$) from the thruster frame into the forces and troques in the body and earth frames, similar to Eq. \ref{eq:allocation_one_thruster}.
However, because an articulated thruster has two force sub-components, $\boldsymbol{T}$ will be a 3 by 2 matrix where the first column of will be $[1\ 0\ 0]^T$ and the second column will be $[0\ 1\ 0]^T$.
Furthermore, two force elements will be added in $\boldsymbol{F}$ to represent an articulated thruster.
Therefore, the size of $\boldsymbol{F}$ and the width of $\boldsymbol{M}_k$ will become $N+2M$ where N and M is the number of fixed thrusters and articulated thrusters, respectively.

Next, we define constraints for each articulated thruster based on its hardware constraints and limitations.
As shown in Fig. \ref{fig:thruster_diagram}, a backward driveable thruster can generate forces within the blue and green fan-shaped areas.
However, a quick reverse on thruster RPM may cause current spikes and backward rotation may have reduced efficiency for asymmetric propellers.
Therefore, we pose a limit such that the thruster can only produce positive thrust.
Without such a limit, $n$ number of articulated thrusters will result in $2^n$ sets of constraints for the QP solver, which becomes less scalable with increased number of articulated thrusters.
After that, we derive other two constraints to limit the force from the articulated thruster inside the green region.
The angle of the green fan is determined based on the servo speed ($\omega$) and the iteration time of the controller ($\delta t$), such that the needed forces is achievable when rotating the thruster.
Base on that we formulated two additional contraints in Eq. \ref{eq:constraints} to bound the force inside the angle of the green region.
The two forces from the articulated thruster will be solved at the same time as other fixed thrusters using Eq.\ref{eq:qp_solver}.
Once the sub-components are known, the actual force for an articulated thruster is computed using Eq. \ref{eq:solution} and the angle from its current value is computed in Eq. \ref{eq:solution2}.

\section{Results}

\begin{figure}[h]
    \centering
    \includegraphics[width=1.0\linewidth,trim=0cm 0cm 0cm 0cm,clip]{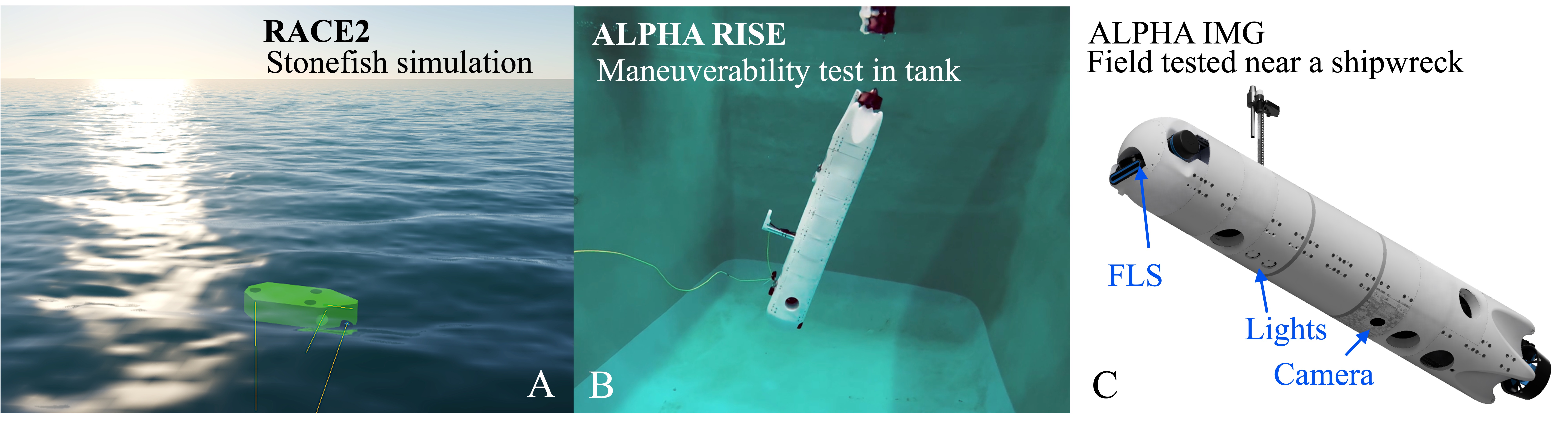}
    \vspace{-4ex}
    \caption{Three AUVs with MVP implemented tested in different environments}
    \label{fig:auv_platforms}
\end{figure}

We have implemented the MVP hardware and software on three different AUVs shown in Fig. \ref{fig:auv_platforms}.
They run the same source code and hardware but with different configuration files, which demonstrate the scalability and customizability of the MVP framework.
Tests have been conducted in simulation, indoor tank, and outdoor to validate their performance.
In Fig. \ref{fig:auv_platforms}.A, the AUV has two fixed thrusters (one vertical thruster and one horizontal thruster) and two articulated thrusters at the stern and was tested in the Stonefish Simulator.
The AUV in Fig. \ref{fig:auv_platforms}.B is equipped with four thrusters (two vertical thrusters, one horizontal thruster, and one main thruster) and we have demonstrated acrobatic movement with MVP in an indoor tank.
Finally, the AUV shown in Fig. \ref{fig:auv_platforms}.C has a similar thruster configuration with the AUV in Fig. \ref{fig:auv_platforms}.B and is equipped with forward-looking sonar and downward-looking camera for seafloor survey applications.
We have deployed the AUV near a shipwreck with MVP running onboard.

\vspace{-1ex}
\vspace{-2ex}

\subsection{Simulation}
\begin{figure}[h]
    \centering
    \includegraphics[width=1.0\linewidth,trim=0cm 1cm 0cm 0cm]{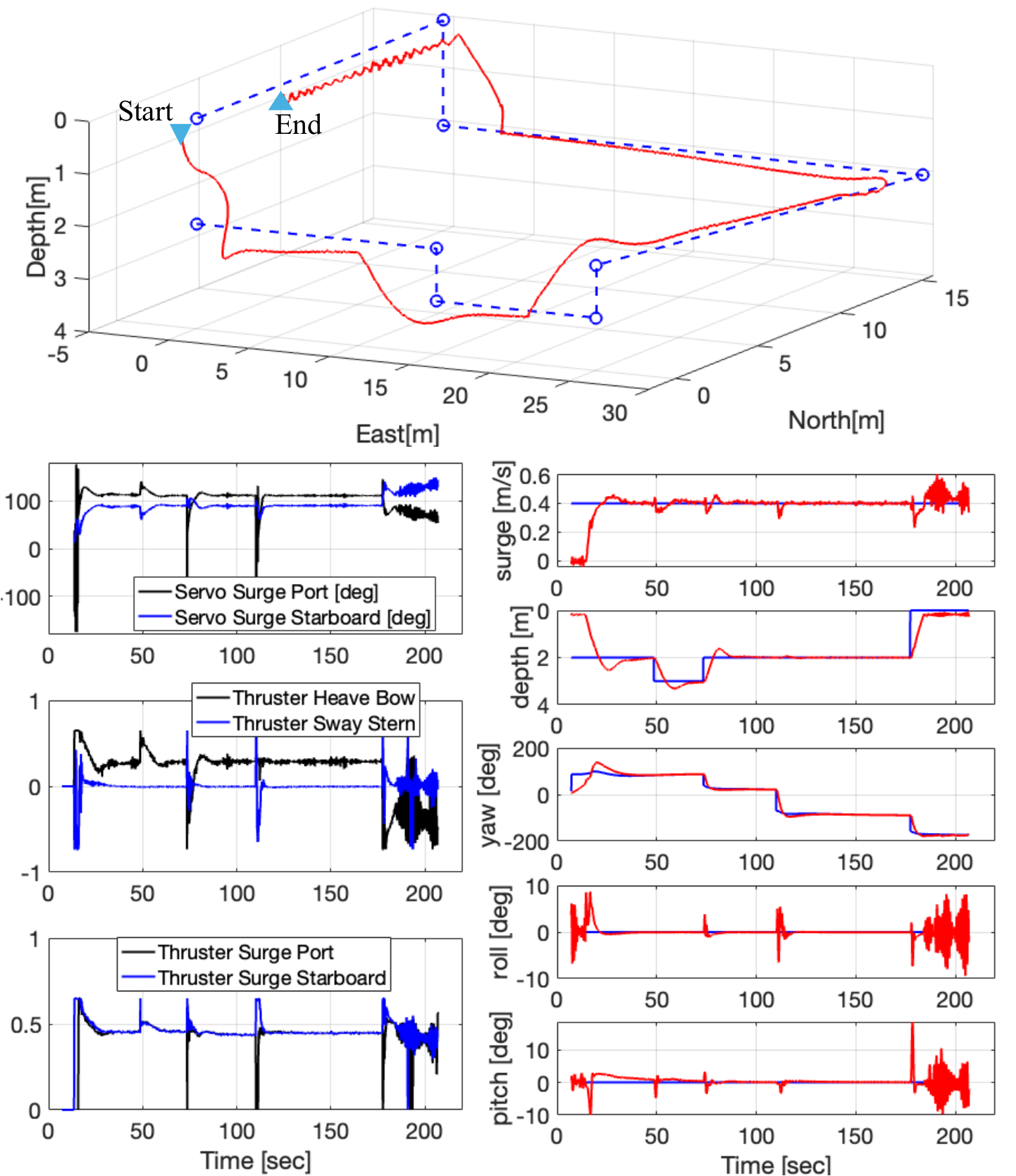}
    \caption{Simulation results for the AUV with dual articulated thrusters. Top: the 3D path of the mission, bottom-left: actuator commands, bottom right: the desired (blue) and current (red) vehicle states.}
    \label{fig:race2_result}
\end{figure}

The AUV equipped shown in Fig. \ref{fig:auv_platforms}.A is tested in the Stonefish simulator with results presented in Fig. \ref{fig:race2_result}.
In the top panel, we show the overall AUV trajectory during a waypoint following mission.
The AUV (red path) has successfully followed 5 waypoints (forming 5 line segments at different depths) shown in blue dash-line.
The bottom left figures depicts how the actuators (thrusters and servos) are controlled. 
We observe spikes in the thruster commands that are mainly due to the sudden changes when transitioning from one line segment to another one.
Towards the end of the mission, we observe high oscillation in thruster commands, which is mainly due to the buoyancy changes when the AUV is cruising at the surface.
From the bottom right panel, we could see the controller resulted in a fast rise time and minimum steady-state error when simultaneously controlling five DOFs (surge, depth, roll, pitch yaw).
Again, the oscillation towards the end of the mission is caused by the buoyancy changes since the desired depth is slightly shallower that the AUV can reach (above the surface).

\vspace{-2ex}

\subsection{Real-world experiments}
The second AUV show in Fig. \ref{fig:auv_platforms} is developed for conducting underwater inspection missions with acrobatic movements.
With the MVP framework integrated, we have performed tank tests to validate the controller and the acrobatic capability. 

\begin{figure}[h]
    \centering
    \includegraphics[width=1.0\linewidth,trim=0cm 0.8cm 0cm 0cm]{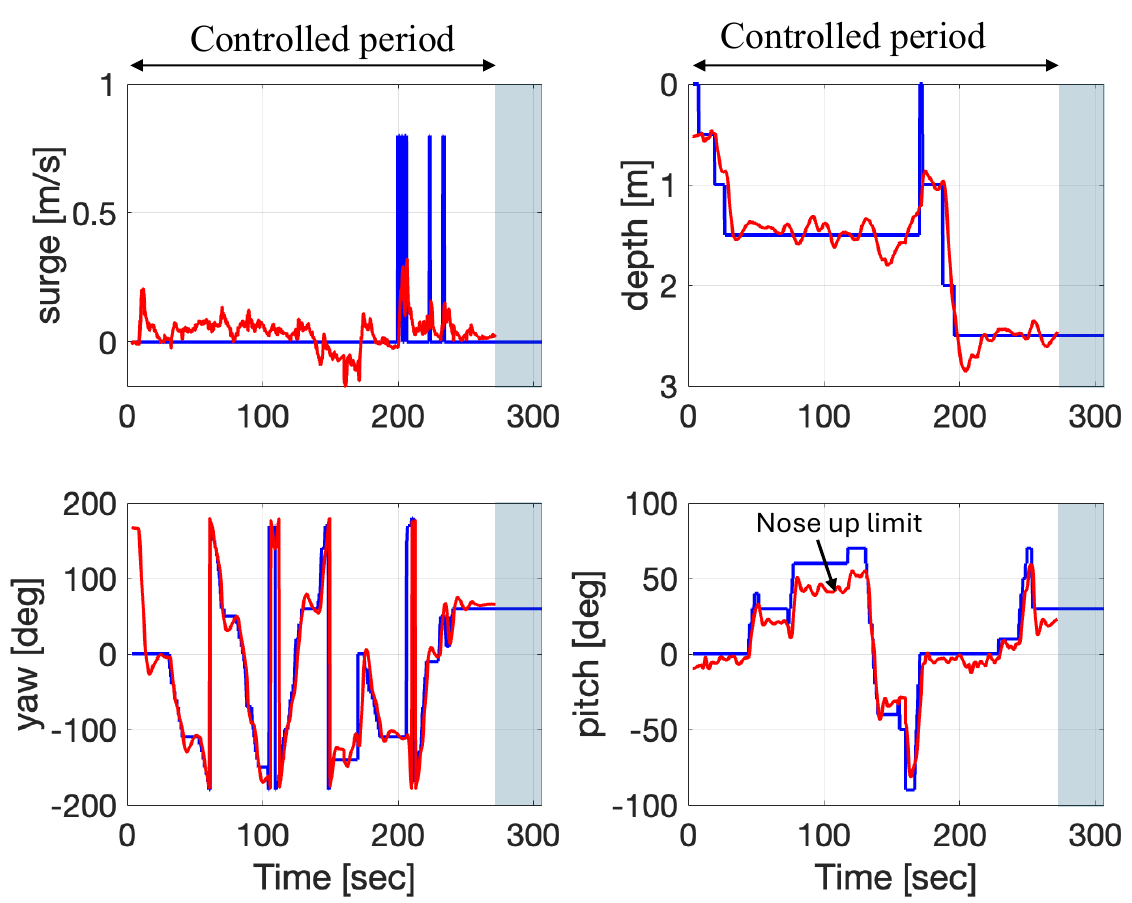}
    \caption{AUV pose changes during the tank tests.}
    \label{fig:rise_tank_data}
\end{figure}

In the tank test, we run the AUV in teleoperation state where the desired pose setpoints (surge, depth, heading and pitch) are altered on a joystick. 
As shown in Fig. \ref{fig:rise_tank_data}, the vehicle tracks the desired pose set points (blue) responsively. 
However, we observed small stead-state error in pitch that may due to the small I-gain for the pitch controller, and the increasing restoring moments at high pitch angles.
Also, the AUV has asymmetrical pitch limits (about 50 when nose up and 80 when nose down).
We set a small P-gain in surge due to the consideration of the tank size, therefore, a slow response time can be observed in the surge plot in Fig. \ref{fig:rise_tank_data}.

\vspace{-2ex}

The third AUV  shown in Fig. \ref{fig:auv_platforms} is developed for underwater seafloor mapping with a suite of perception sensors, including forward-looking imaging sonar (Blueprint Subsea Oculus) and downward-looking camera (DeepWater Exploration stellarHD Machine Vision camera).
We have conducted field tests with the AUV using MVP framework to survey the seabed near a shipwreck (about 5 meters deep) in Narragansett Bay.

\vspace{-1ex}
\begin{figure}[h]
    \centering
    \includegraphics[width=1.0\linewidth,trim=0cm 0cm 3cm 0cm,clip]{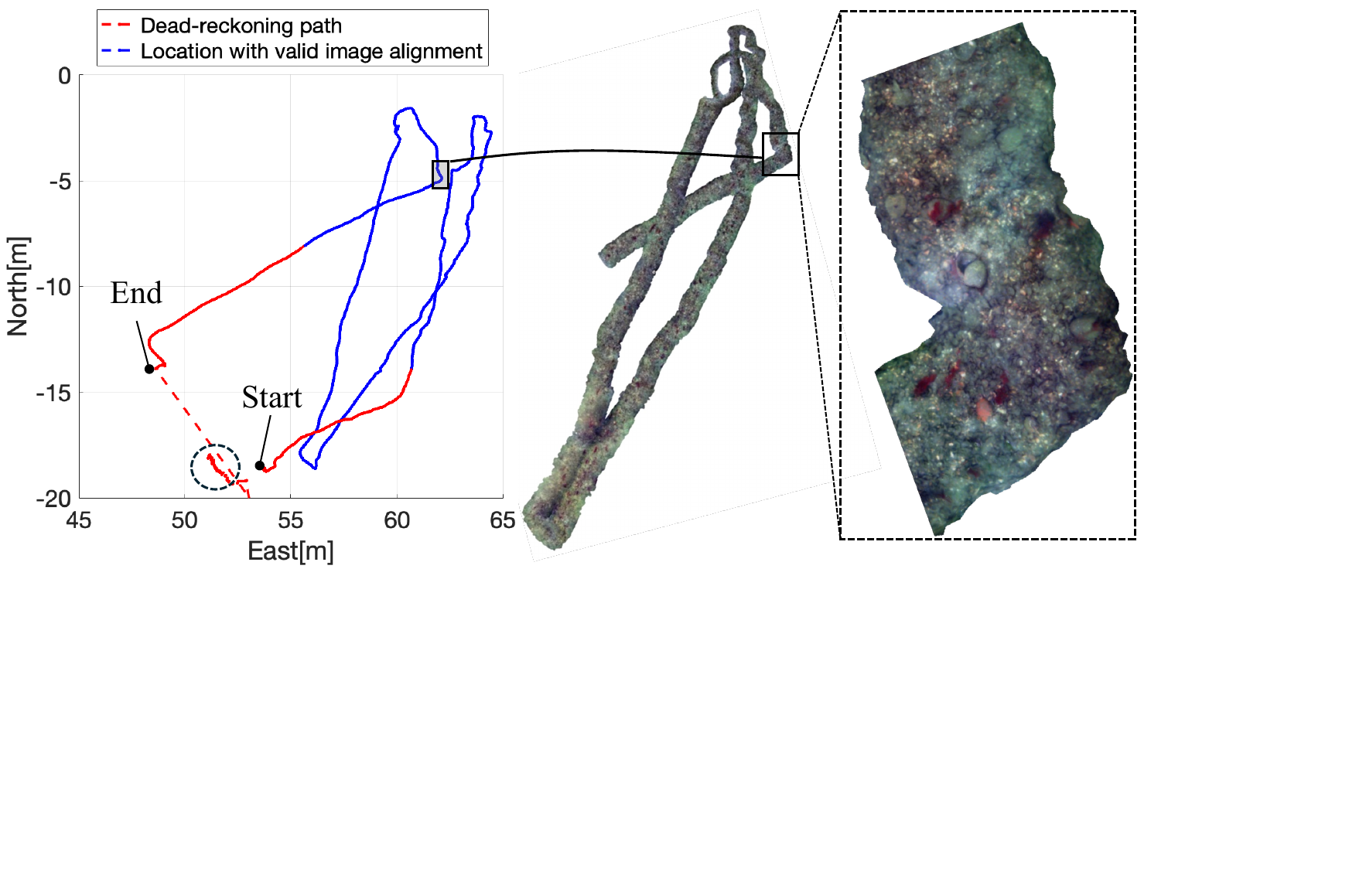}
    \vspace{-20ex}
    \caption{An overview of the mission. Left: AUV path in 2D, middle: photo moscaing results, right: zoom-in view of the highlighted section.}
    \label{fig:alpha_img_render}
\end{figure}

Figure \ref{fig:alpha_img_render} presents the overall AUV path and the moscaing result (generated in Metashape) using the camera images.
The camera images are color-corrected using the script from \cite{matlab_code} based on methods presented in \cite{Ancuti2018ColorBA}.
Compared the rendering results with the dead-reckoning path, we found the accumulated drift exists in the dead-reckoning path.

\begin{figure}[h]
    \centering
    \includegraphics[width=0.8\linewidth,trim=0cm 1.5cm 1cm 0cm]{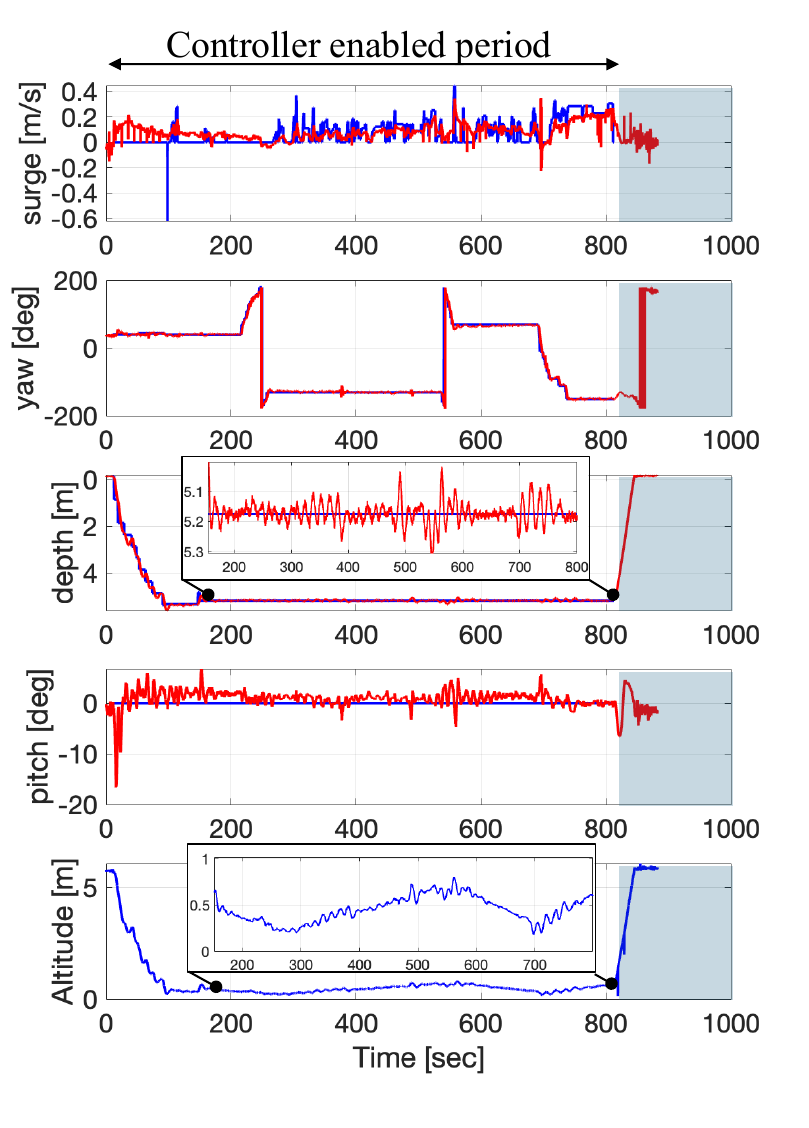}
    \caption{AUV pose changes during the low-altitude visual survey.}
    \label{fig:alpha_img_result}
\end{figure}

In Fig. \ref{fig:alpha_img_result} we present the AUV's motion data during this dive.
The heading control of the AUV is effective that can track the desired heading commands in a relative short time.
In Fig. \ref{fig:alpha_img_result}, we also demonstrate the excellent depth holding capability.
During the survey, the depth is controlled within 0.1 m depth band at an altitude (measured by the DVL) of about 0.5 m.
As shown in the surge plot, we observe non-zero surge during diving.
We believe this is mainly due to the non-zero pitch.
In order to keep the AUV underwater, the vertical thrusters will generate downward thrusts which will lead to a forward motion if the AUV is pitching up.
When we speed up the AUV during the later potion of the dive, we see reduced pitch and the AUV intended to maintain the desired surge.

\vspace{-3ex}
\vspace{-3ex}

\section{Conclusion and future works}
In this paper, we have presented a new framework, called MVP, towards modular and accessible AUV system development.
The framework has hardware and software components that are openly accessible through the GitHub repositories \cite{mvp_readme}.
The hardware provides a set of PCBs that integrates different interface functionalities for underwater thrusters, sensors and payloads, while the software provides an all-in-one solution for AUV guidance navigation and control, operation, and simulation interfaces.
Particularly in this paper, we have introduced our new controller feature that is compatible with articulated thrusters.
The presented method allows solving forces and angles for articulated thrusters in an effective way.
To demonstrate the system compatibility, we have implemented MVP on three different AUV platforms, and we have presented results from simulation, indoor tanks, and outdoor experiments.
The excellent control capability allowed us to conduct a visual survey of the seabed near a shipwreck using AUVs.
During the survey, the depth was kept within 0.1 m depth band, and the images from the downward-looking camera are rendered into a mosaic in Metashape with detailed texture information showing objects on the seabed, such as shell deposits and vegetation.

Several high-level MVP software modules (i.e., mvp\_autonomy and mvp\_acomm) are still under development, and the SBC PCB is currently under revision.
The upgrades will be installed on the AUV platforms and further validated in the field.
The ROS2 version of the MVP software is currently under beta testing.
We expect to provide a stable release in December 2024, which will be compatible with the latest ROS (Jazzy) and Ubuntu 24.04.

\bibliographystyle{IEEEtran}
\bibliography{IEEEabrv,bibliography}

\begin{thebibliography}{10}
\providecommand{\url}[1]{#1}
\csname url@samestyle\endcsname
\providecommand{\newblock}{\relax}
\providecommand{\bibinfo}[2]{#2}
\providecommand{\BIBentrySTDinterwordspacing}{\spaceskip=0pt\relax}
\providecommand{\BIBentryALTinterwordstretchfactor}{4}
\providecommand{\BIBentryALTinterwordspacing}{\spaceskip=\fontdimen2\font plus
\BIBentryALTinterwordstretchfactor\fontdimen3\font minus \fontdimen4\font\relax}
\providecommand{\BIBforeignlanguage}[2]{{%
\expandafter\ifx\csname l@#1\endcsname\relax
\typeout{** WARNING: IEEEtran.bst: No hyphenation pattern has been}%
\typeout{** loaded for the language `#1'. Using the pattern for}%
\typeout{** the default language instead.}%
\else
\language=\csname l@#1\endcsname
\fi
#2}}
\providecommand{\BIBdecl}{\relax}
\BIBdecl

\bibitem{Edge2020DesignAE}
\BIBentryALTinterwordspacing
C.~Edge, S.~S. Enan, M.~Fulton, J.~Hong, J.~Mo, K.~Barthelemy, H.~Bashaw, B.~Kallevig, C.~Knutson, K.~Orpen, and J.~Sattar, ``Design and experiments with loco auv: A low cost open-source autonomous underwater vehicle*,'' \emph{2020 IEEE/RSJ International Conference on Intelligent Robots and Systems (IROS)}, pp. 1761--1768, 2020. [Online]. Available: \url{https://api.semanticscholar.org/CorpusID:214606022}
\BIBentrySTDinterwordspacing

\bibitem{Duecker2020HippoCampusXA}
\BIBentryALTinterwordspacing
D.~A. Duecker, N.~Bauschmann, T.~Hansen, E.~J. Kreuzer, and R.~Seifried, ``Hippocampusx – a hydrobatic open-source micro auv for confined environments,'' \emph{2020 IEEE/OES Autonomous Underwater Vehicles Symposium (AUV)(50043)}, pp. 1--6, 2020. [Online]. Available: \url{https://api.semanticscholar.org/CorpusID:227277895}
\BIBentrySTDinterwordspacing

\bibitem{Griffiths2016AVEXISAquaVE}
\BIBentryALTinterwordspacing
A.~Griffiths, A.~Dikarev, P.~R. Green, B.~Lennox, X.~Poteau, and S.~Watson, ``Avexis—aqua vehicle explorer for in-situ sensing,'' \emph{IEEE Robotics and Automation Letters}, vol.~1, pp. 282--287, 2016. [Online]. Available: \url{https://api.semanticscholar.org/CorpusID:8898286}
\BIBentrySTDinterwordspacing

\bibitem{Rypkema2023PerseusAT}
\BIBentryALTinterwordspacing
N.~R. Rypkema, S.~Randeni, M.~Sacarny, M.~Benjamin, and M.~Triantafyllou, ``Perseus auv: Towards linear convoying of agile a-sized auvs through acoustic track-and-trail,'' \emph{2023 IEEE/RSJ International Conference on Intelligent Robots and Systems (IROS)}, pp. 6206--6213, 2023. [Online]. Available: \url{https://api.semanticscholar.org/CorpusID:266195823}
\BIBentrySTDinterwordspacing

\bibitem{Randeni2022MorpheusAA}
\BIBentryALTinterwordspacing
S.~Randeni, M.~Sacarny, M.~B. Benjamin, and M.~S. Triantafyllou, ``Morpheus: An a-sized auv with morphing fins and algorithms for agile maneuvering,'' \emph{ArXiv}, vol. abs/2212.11692, 2022. [Online]. Available: \url{https://api.semanticscholar.org/CorpusID:254974201}
\BIBentrySTDinterwordspacing

\bibitem{Benjamin2010NestedAF}
\BIBentryALTinterwordspacing
M.~R. Benjamin, H.~Schmidt, P.~Newman, and J.~J. Leonard, ``Nested autonomy for unmanned marine vehicles with moos‐ivp,'' \emph{Journal of Field Robotics}, vol.~27, 2010. [Online]. Available: \url{https://api.semanticscholar.org/CorpusID:14713090}
\BIBentrySTDinterwordspacing

\bibitem{Palomeras2012COLA2AC}
\BIBentryALTinterwordspacing
N.~Palomeras, A.~El-Fakdi, M.~Carreras, and P.~Ridao, ``Cola2: A control architecture for auvs,'' \emph{IEEE Journal of Oceanic Engineering}, vol.~37, pp. 695--716, 2012. [Online]. Available: \url{https://api.semanticscholar.org/CorpusID:30867076}
\BIBentrySTDinterwordspacing

\bibitem{Gezer2022WorkingTT}
E.~C. Gezer, M.~Zhou, L.~Zhao, and W.~McConnell, ``Working toward the development of a generic marine vehicle framework: Ros-mvp,'' \emph{OCEANS 2022, Hampton Roads}, pp. 1--5, 2022.

\bibitem{mvp_readme}
\BIBentryALTinterwordspacing
``{MVP Readme}.'' [Online]. Available: \url{https://github.com/uri-ocean-robotics/mvp\_readme}
\BIBentrySTDinterwordspacing

\bibitem{Cielak2019StonefishAA}
\BIBentryALTinterwordspacing
P.~Cieślak, ``Stonefish: An advanced open-source simulation tool designed for marine robotics, with a ros interface,'' \emph{OCEANS 2019 - Marseille}, pp. 1--6, 2019. [Online]. Available: \url{https://api.semanticscholar.org/CorpusID:204701708}
\BIBentrySTDinterwordspacing

\bibitem{MooreStouchKeneralizedEkf2014}
T.~Moore and D.~Stouch, ``A generalized extended kalman filter implementation for the robot operating system,'' in \emph{Proceedings of the 13th International Conference on Intelligent Autonomous Systems (IAS-13)}.\hskip 1em plus 0.5em minus 0.4em\relax Springer, July 2014.

\bibitem{goby}
\BIBentryALTinterwordspacing
``Goby3.'' [Online]. Available: \url{https://goby.software/3.0/}
\BIBentrySTDinterwordspacing

\bibitem{Fossen2006ASO}
\BIBentryALTinterwordspacing
T.~I. Fossen and T.~A. Johansen, ``A survey of control allocation methods for ships and underwater vehicles,'' \emph{2006 14th Mediterranean Conference on Control and Automation}, pp. 1--6, 2006. [Online]. Available: \url{https://api.semanticscholar.org/CorpusID:14665084}
\BIBentrySTDinterwordspacing

\bibitem{Jin2015SixDegreeofFreedomHC}
\BIBentryALTinterwordspacing
S.~Jin, J.~Kim, J.~Kim, and T.~Seo, ``Six-degree-of-freedom hovering control of an underwater robotic platform with four tilting thrusters via selective switching control,'' \emph{IEEE/ASME Transactions on Mechatronics}, vol.~20, pp. 2370--2378, 2015. [Online]. Available: \url{https://api.semanticscholar.org/CorpusID:23993841}
\BIBentrySTDinterwordspacing

\bibitem{matlab_code}
\BIBentryALTinterwordspacing
``Color balance and fusion for underwater image enhancement,'' \emph{MATLAB Central File Exchange}, 2024. [Online]. Available: \url{https://www.mathworks.com/matlabcentral/fileexchange/120473-color-balance-and-fusion-for-underwater-image-enhancement}
\BIBentrySTDinterwordspacing

\bibitem{Ancuti2018ColorBA}
\BIBentryALTinterwordspacing
C.~O. Ancuti, C.~Ancuti, C.~D. Vleeschouwer, and P.~Bekaert, ``Color balance and fusion for underwater image enhancement,'' \emph{IEEE Transactions on Image Processing}, vol.~27, pp. 379--393, 2018. [Online]. Available: \url{https://api.semanticscholar.org/CorpusID:9258737}
\BIBentrySTDinterwordspacing

\end{thebibliography}

\end{document}